\documentclass[letterpaper, 10 pt, conference]{ieeeconf}  

\IEEEoverridecommandlockouts                              

\usepackage{graphics} 
\usepackage{epsfig} 
\usepackage{mathptmx} 
\usepackage{times} 
\usepackage{amsmath} 
\usepackage{amssymb}  
\usepackage[mathscr]{eucal}
\usepackage{subcaption}
\usepackage{booktabs}
\usepackage{hyperref}
\usepackage{algorithm}
\usepackage{algorithmic}

\DeclareMathOperator*{\argmax}{argmax}

\title{\LARGE \bf
Continual Learning for Autonomous Robots: A Prototype-based Approach
}

\author{Elvin Hajizada$^{1,2}$, Balachandran Swaminathan$^{3}$, Yulia Sandamirskaya$^{4}$
\thanks{*This work was supported by Intel}%
\thanks{*This work has been submitted to the IEEE for possible publication. Copyright may be transferred without notice, after which this version may no longer be accessible}%
\thanks{$^{1}$Neuromorphic Computing Lab, Intel Labs, Munich, Germany}
\thanks{$^{2}$School of Computation, Information and Technology, Technical University of Munich, Germany
        {\tt\small elvin.hajizada@tum.de}}%
\thanks{$^{3}$Pennsylvania State University, 
        {\tt\small bus44@psu.edu}}%
\thanks{$^{4}$Institute of Computational Life Sciences (ICLS), Zurich University of Applied Sciences (ZHAW), Schloss 1, 8820 Wädenswil, Switzerland}
}

\begin{document}

\maketitle
\thispagestyle{empty}
\pagestyle{empty}

\begin{abstract}

Humans and animals learn throughout their lives from limited amounts of sensed data, both with and without supervision. Autonomous, intelligent robots of the future are often expected to do the same. The existing continual learning (CL) methods are usually not directly applicable to robotic settings: they typically require buffering and a balanced replay of training data. A few-shot online continual learning (FS-OCL) setting has been proposed to address more realistic scenarios where robots must learn from a non-repeated sparse data stream. To enable truly autonomous life-long learning, an additional challenge of detecting novelties and learning new items without supervision needs to be addressed. We address this challenge with our new prototype-based approach called Continually Learning Prototypes (CLP). In addition to being capable of FS-OCL learning, CLP also detects novel objects and learns them without supervision. To mitigate forgetting, CLP utilizes a novel metaplasticity mechanism that adapts the learning rate individually per prototype. CLP is rehearsal-free, hence does not require a memory buffer, and is compatible with neuromorphic hardware, characterized by ultra-low power consumption, real-time processing abilities, and on-chip learning. Indeed, we have open-sourced a simple version of CLP in the neuromorphic software framework Lava, targetting Intel's neuromorphic chip Loihi 2. We evaluate CLP on a robotic vision dataset, OpenLORIS. In a low-instance FS-OCL scenario, CLP shows state-of-the-art results. In the open world, CLP detects novelties with superior precision and recall and learns features of the detected novel classes without supervision, achieving a strong baseline of $99\%$ base class and $65\%/76\%$ (5-shot/10-shot) novel class accuracy.
\end{abstract}

\section{Introduction}
\label{sec:intro}

Autonomous, interactive, and lifelong learning are features of human intelligence that distinguish it from the machine intelligence of the modern age. Current machine learning methods outperform both humans and hand-crafted algorithm on a given static data set, but fail spectacularly when the key assumptions of the neural network training scheme, e.g., of identically and independently distributed (i.i.d.) data, are violated~\cite{Hadsell2020CLreview}. To address the limitations of static data distribution,  Continual Learning (CL) is an emerging topic in AI. The main issue CL aims to address is catastrophic forgetting, the phenomenon that reflects the trade-off between attaining new knowledge while retaining the old knowledge, also known as the plasticity-stability dilemma~\cite{Yan2021der}. Replay, regularization, parameter isolation, and network expansion methods have been some of the most common techniques in the CL literature~\cite{Hadsell2020CLreview, Lange2022CLreview}. 

\begin{figure}[t]
  \centering
   \includegraphics[width=1\linewidth]{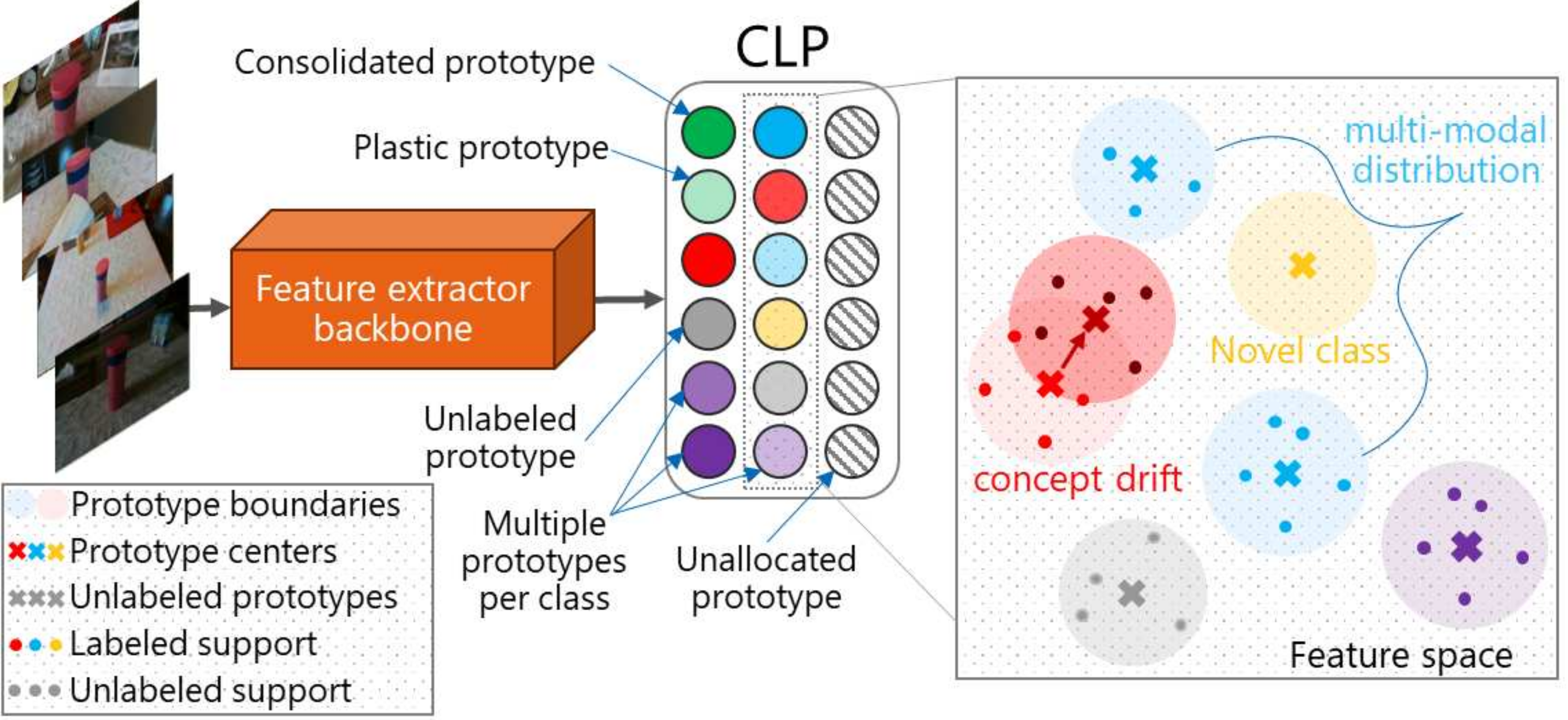}
   \caption{Continually Learning Prototypes: overview}
   \label{fig:clp_diagram}
   \vspace{-15pt}
\end{figure}

However,  the problem of catastrophic forgetting is not the only challenge that must be addressed to close the gap between today's training of deep neural networks and the more natural learning processes we know from humans and animals. Recently, learning objects from a few labeled samples provided through a non-repeating stream of input has gained attention~\cite{hayes2022online, michieli2023fsocl}. This setting, formally called few-shot online continual learning (FS-OCL), is a step towards realistic learning for robots.

Yet, FS-OCL is still far from real-world learning scenarios. For instance, human concept learning involves not only a small amount of direct instruction (e.g., parental labeling) but also large amounts of unlabeled experience (e.g., observation of objects without naming them). This unsupervised learning is continual, autonomous, and interactive. A strong driver of this learning process is novelty detection -- the ability to recognize an instance as something not seen before~\cite{Bendale2015nno, roady2020ood}. On the contrary, a common close-world assumption in deep learning is that all the test instances are from the learned classes. Detecting novel instances alone is not enough, however, as such a system should also integrate these novelties into its knowledge, even without supervision~\cite{jafarzadeh2020openWorldReview}. Therefore, we extend FS-OCL to include open-world and semi-supervised learning to achieve the most natural continual object learning setting for robots, which we shortly call \textbf{O}pen \textbf{W}orld \textbf{C}ontinual \textbf{L}earning (OWCL). 

We propose \textit{\textbf{Continually Learning Prototypes (CLP)}}, a comprehensive learning algorithm to tackle OWCL, which is capable of online continual learning from few-shot (un)labeled data in the open world with unknowns. 
We introduce a novel dynamic adaptation mechanism for the learning rate of individual prototype neurons (metaplasticity~\cite{jedlicka2022metaplasticity}) to address the stability-plasticity dilemma and, hence, catastrophic forgetting. 
Furthermore, we allow multiple prototypes per class, which can be allocated over time via novelty detection and dynamically adapt in a semi-supervised manner from a stream of input. 
Crucially, CLP is rehearsal-free and does not maintain a memory buffer, as it targets robotic platforms, which generally have compute, memory, and energy constraints. 
To further this argument, we also implemented a simpler version of CLP for neuromorphic chip Loihi 2 and open-sourced it as part of the Lava software framework\footnote{Visit \href{https://github.com/lava-nc/lava/tree/main/tutorials/in_depth/clp}{CLP in Lava}.}. 
Note that the details and results of the neuromorphic implementation are beyond the scope of this paper.
Our contributions are summarized as follows:
\begin{itemize}
    \item We tackle a novel learning scenario called open-world continual learning (OWCL) to evaluate robotic object learning in the most realistic way. This scenario assumes data becomes available sample-by-sample in open-world, where novel classes may appear spontaneously with or without labels. These instances need to be detected and learned, possibly with few shots, all the while avoiding catastrophic forgetting.
    \item  This is combined with positive or negative updates to winner prototypes based on the label. If the label is not available, CLP behaves as a novelty detection-assisted clustering algorithm that detects and learns novel instances while being able to follow gradual concept drifts.
    \item We dynamically adjust multiple prototypes per class in a semi-supervised manner from streaming input. 
\end{itemize}

\section{Related work}
\label{sec:related_work}
\textbf{Continual Learning.} Continual Learning (CL) research has predominantly addressed the challenges of learning from a sequence of tasks without catastrophic forgetting. The most common techniques to achieve CL are replay, regularization, parameter isolation, and network expansion. Replay methods generally utilize a memory buffer that retains representative examples from learned data distribution~\cite{rebuffi2017icarl,de2021cope}. Alternatively, synthetic memories can be generated using a generative model and replayed to the main model~\cite{shin2017dgr}. In both cases, the goal is to mimic i.i.d training by facilitating co-optimizing old and new class learning. On the other hand, popular regularization techniques~\cite{chaudhry2018agem} use the stored samples to constrain the new task updates in the network. One approach is to estimate the importance of each learned weight and use this information to limit the changes applied to the important weights, hence preserving the knowledge~\cite{kirkpatrick2017ewc, aljundi2018mas, zhang2020dmc}. Finally, the third category encapsulates parameter isolation where task-specific parameters are isolated;  they stay unchanged while learning the other tasks~\cite{rusu2016pnn, mallya2018piggyback}. These methods can result in dynamically growing networks as they learn~\cite{rusu2016pnn} or have a fixed network size but dedicate parts to each task~\cite{mallya2018packnet, mallya2018piggyback}. 
While these techniques cumulatively address catastrophic forgetting, some make strong i.i.d assumptions or assume the availability of labeled samples arriving in batches. This makes it unsuitable for the autonomous and open-world online continual learning scenarios we consider in this work. 

\textbf{Prototype-based Learning.} For memory and compute-constrained embedded devices like robots, storing representative examples or ``prototypes" of seen classes has proven beneficial for learning in non-stationary environments. Prototype-based learning methods like self-organizing map (SOM), learning vector quantization (LVQ), initially introduced by Kohonen~\cite{kohonen1990som}, laid the foundation for extensive adaption in incremental learning literature of the pre-deep learning era~\cite{losing2018incrLearningReview, xu2012ilvq, beyer2013dyng}. 
Recently, deep learning-based CL methods began incorporating prototypes in the network's final layer for inference and/or learning~\cite{parisi2018prototypeSequence,zhu2021protoAug, de2021cope, hayes2020deepSLDA}, or as memory buffers for pattern replay~\cite{hayes2019protoReplay}. Additionally, these techniques have been applied in few-shot class incremental learning (FSCIL), where networks learn incrementally from small batches of labeled samples~\cite{snell2017protoNet, hersche2022constrained}. However, the idea that new classes are presented in regular, iterable batches is impractical, and the common practice of using a single prototype per class overlooks the complexity of real-world class distributions~\cite{allen2019imp, ayub2020fscilIROS}, which may require multiple prototypes.
More recently, CBCL~\cite{ayub2020cbcl} proposed novelty detection-assisted clustering that produces multiple prototypes per class. Like prior ones, this method presupposes labeled samples and typically calculates prototypes as the label-based (running) average of cluster samples, employing a linearly decaying learning rate without error-based prototype updates. Conversely, original prototype-based supervised approaches~\cite{Shen2020online_semi_lvq, beyer2013dyng,xu2012ilvq} dynamically adjust prototypes via stochastic gradient descent. We expand upon all these previous works by introducing a dynamic and independent plasticity mechanism (metaplasticity), akin to a learning rate for each prototype, to address the stability-plasticity dilemma. Furthermore, we introduce the capability of semi-supervised learning while still allowing more than one prototype per class. Inspired by LVQ methods~\cite{losing2018incrLearningReview, Shen2020online_semi_lvq, beyer2013dyng}, we update prototypes also when they make errors. These enhancements allow us to effectively manage the evolving demands of open-world continual learning scenarios.

\textbf{Open World Learning.} One such demand is when the robot operates in an open world where novel objects appear all the time. Therefore, novelty detection is also necessary for realistic continual learning. There is a significant body of work on open-set and open-world recognition~\cite{Bendale2015nno, bohus2022wild, Ren2021wander, jafarzadeh2020openWorldReview} that demonstrates continuous detection of novel classes. Subsequently, OWCL, as a truly realistic and natural learning setting for an autonomous robot, encompasses all of 1) online continual learning, 2) few-shot learning, 3) open-set recognition, and 4) semi-supervised learning of new and old categories. To the best of our knowledge, a comprehensive algorithm that combines all such learning scenarios faced by autonomous agents is heavily under-explored in the literature~\cite{jafarzadeh2020openWorldReview}. 

This work aims to build a flexible algorithm for continual online learning adaptable to semi-supervised open-world contexts with the help of prototypes.
\section{Problem Setting}
\label{sec:problem}

\textbf{Few-shot Online Continual Learning (FS-OCL).} In the FS-OCL setting~\cite{michieli2023fsocl}, a model sequentially processes and learns a dataset  $\mathscr{D}={(\boldsymbol{x_t}, y_t)}^{\infty}_{t=1}$ one instance at a time, where $(\boldsymbol{x_t}, y_t)$ is an example-label pair received at time $t$, with $\boldsymbol{x_t} \in \mathbb{R}^D$. Furthermore, only a few such samples can be provided, as it is time-consuming and tedious for users. The model can be decomposed into a feature extractor and an online classiﬁer: $f(\boldsymbol{x_t}) = F(G(\boldsymbol{x_t}))$, where $G(\cdot):\mathbb{R}^D \rightarrow \mathbb{R}^d$ is a backbone, $F(\cdot)$ is a classifier and $d$ is the dimension of the feature vector. An autonomous learning agent would not have control over the order of data it receives, meaning there is no guarantee of i.i.d. in relation to previous samples. In addition, the model assumes that each sample is seen only once and cannot be stored. The agent must act on the newly learned knowledge without delay and seamlessly arbitrate between learning and inference. Besides, no task labels or boundaries are available to the model. 

\textbf{Open-world Continual Learning.} We extend the FS-OCL protocol to include other crucial aspects of real-world autonomous learning. First, at any time t, we consider the set of known object classes  $\mathscr{K} = {1, 2,.., C} \in \mathbb{N}_+$ where $\mathbb{N}_+$ denotes the set of positive integers. We embrace the ``open-world" assumption in our simulation of the natural learning environment for autonomous agents, acknowledging the existence of an unknown set of classes $\mathscr{U} = {C+1,...}$, that might be encountered later. The known object classes $\mathscr{K}$ are regarded as base classes and labeled in the dataset, ensuring sufficient samples exist for each class to facilitate effective learning. Conversely, unknown classes emerge as the open-world scenario progresses, presenting a limited number of unlabeled samples. At this stage, the established model $\mathscr{M}_C$, trained on $C$ classes, is tasked with identifying and learning these new instances in a few-shot manner, operating without supervision. The model aims to acquire new knowledge while preserving existing foundational knowledge. Consistent with the principles of semi-supervised continual learning~\cite{Shen2020online_semi_lvq, Joseph2021openDetect}, it is assumed that real labels will intermittently be provided for the unsupervised gained knowledge.

\section{Continually Learning Prototypes}
\label{sec:clp}

To tackle the challenges of Open-World Continual Learning (OWCL) for autonomous agents, we propose a novel method called \textit{Continually Learning Prototype} (CLP), an online open-world continual learning method. This method is designed for online learning in dynamic real-world environments, offering several key capabilities: 1) the ability to learn from a continuous stream of data without experiencing catastrophic forgetting, 2) learning in situations with limited data (few-shot learning), 3) detecting and adapting to novel information, and 4) learning without the need for supervision. CLP is built as a prototype-based algorithm, utilizing prototypes as representatives for clusters of instances~(Fig.~\ref{fig:clp_diagram}), and it seamlessly enables online, few-shot learning~(Sec.~\ref{sec:prototype_met}). Crucially, it introduces a novel mechanism for adapting the learning rate of individual prototypes~(Sec.~\ref{sec:meta_plasticity}), addressing the plasticity-stability dilemma and hence catastrophic forgetting, inspired by the metaplasticity observed in biological neurons~\cite{jedlicka2022metaplasticity, Kudithipudi2022bioCL}. Furthermore, CLP pursues the approach of on-demand allocation of new prototypes and incrementally constructing a multi-modal (multi-cluster) representation for each class, efficiently capturing both simple and complex classes~(Sec.~\ref{sec:multimodal}). The allocation process is triggered by CLP's novelty detection mechanism, crucial tackling open-world scenarios~(Sec.~\ref{sec:openWorld}). 

\subsection{Prototype-based Learning}\label{sec:prototype_met}
At its core, our classifier method (CLP) employs the concept of prototypes. It determines the winner prototype that is most similar to the input feature vector and updates it towards or away from this vector, based on whether it made a correct or incorrect prediction, respectively. Most prototype-based methods use Euclidean distance to find the best matching prototype~\cite{Biehl2016}. On the other hand, CLP uses dot-product similarity, together with normalized feature and prototype vectors, as a proxy to achieve cosine similarity, which is shown to be a more powerful similarity measure for large dimensional vectors~\cite{hou2019unifiedClassifier,hersche2022constrained}. The decision to avoid using cosine similarity directly stems from the increased complexity it introduces to the update rule for prototypes. In contrast, the dot product is chosen due to its widespread availability and well-optimized computations in neural accelerators such as GPUs and neuromorphic chips~\cite{Davies2018, hajizada2022interactive}. Therefore, the similarity measure used by CLP can be expressed as follows:
\begin{equation} \label{eq:sim_measure}
s(\boldsymbol{\mu},\boldsymbol{x}) = \boldsymbol{\mu} \cdot \boldsymbol{x},
\end{equation}
where $s(\boldsymbol{\mu},\boldsymbol{x})$ is the similarity between a prototype and input sample $\boldsymbol{\mu},\boldsymbol{x}  \in \mathbb{R}^d$ and $\|\boldsymbol{\mu}\| = \|\boldsymbol{x}\| = 1$. 

\paragraph{Online Learning with Prototypes}
Utilizing the defined similarity measure as the cornerstone, CLP adopts a learning rule rooted in stochastic gradient descent, making it adaptable to online learning settings. This characteristic allows CLP to efficiently handle dynamic learning scenarios. Now, let's define a prototype population as $P = \{(\boldsymbol{\mu^i},l^i)\}_{i=1}^{n}$, where $\boldsymbol{\mu^i},l^i$ are the center and label of a prototype $p_i$. The learning rule to update the winner prototype's mean $\boldsymbol{\mu^*}$ when learning input pair $(x,\hat{y})$ can then be written using the gradient of the similarity measure described in (\ref{eq:sim_measure}):
\begin{equation}
\boldsymbol{\mu^*} \leftarrow \boldsymbol{\mu^*} + \alpha \Psi(y^*,\hat{y}) \nabla_{\boldsymbol{\mu}} s(\boldsymbol{\mu^*},\boldsymbol{x}).\nonumber
\end{equation}
We can simplify this equation by using the equality $\nabla_{\boldsymbol{\mu}} s(\boldsymbol{\mu},\boldsymbol{x})=\nabla_{\boldsymbol{\mu}} s(\boldsymbol{\mu} \cdot \boldsymbol{x})=\boldsymbol{x}$:
\begin{equation} \label{eq:lvq_update}
\boldsymbol{\mu^*} \leftarrow \boldsymbol{\mu^*} + \alpha \Psi(y^*,\hat{y}) \boldsymbol{x}, 
\end{equation}
\begin{equation}
\text{where }\boldsymbol{\mu^*} = \argmax_{i \in [1,n] } \, s(\boldsymbol{\mu^i},\boldsymbol{x}); \hspace{0.5cm} y^* \leftarrow l^*, \nonumber
\end{equation}
\begin{equation}
\Psi(y^*,\hat{y}) =  
\begin{cases}
  +1 & \text{if } y^*=\hat{y} \\
  -1 & \text{otherwise},
\end{cases}
\end{equation}
and $\alpha$ is the learning rate, $y^*$ is the model prediction, which is the label ($l^*$) of the winner prototype, while $\Psi(y^*,\hat{y})$ is the evaluator of the prediction. 

\paragraph{Few-shot Learning with Prototypes}
In addition to learning online, CLP can also swiftly assimilate novel examples into its knowledge without catastrophic forgetting. This is particularly significant in the context of few-shot learning, where the primary challenge is to acquire meaningful insights from limited data without succumbing to overfitting. CLP tackles this challenge by positing a straightforward inductive bias: the existence of an embedding where points group around prototypes, learnable even from scarce data~\cite{snell2017protoNet}. This allows CLP to perform various learning types from its repertoire, not only in scenarios with abundant data but also in one-shot or few-shot settings.

\subsection{Metaplasticity for Continual Learning} \label{sec:meta_plasticity}
While the aforementioned capabilities of CLP make a good start towards our goal of autonomous continual learning, the key feature and contribution of CLP lies in how it manages continual learning. We approach continual learning and catastrophic forgetting from the perspective of the plasticity-stability dilemma. We argue that different parts of knowledge should have different levels of stability, which can be derived from their performance history~\cite{hamker200individualLR,jedlicka2022metaplasticity}. For each piece of continually learned knowledge, i.e., for every prototype, CLP maintains an individual plasticity level or learning rate. The performance history is tracked in another variable called ``goodness". As a prototype evolves and makes correct predictions, its ``goodness" level increases; thus, its learning rate decreases. This is a type of consolidation process that protects knowledge from interference and forgetting during the subsequent learning sessions. However, if a prototype frequently makes erroneous decisions, the opposite happens, and the learning rate starts to increase. Then, the increased plasticity allows this prototype to adjust and fix the faulty behavior. CLP's adaptive learning rate mechanism (metaplasticity) addresses the plasticity-stability dilemma by selectively consolidating useful knowledge while justifiably forgetting outdated or erroneous parts. Hence, we extend the definition of the prototype population to $P = \{(\boldsymbol{\mu^i},\alpha^i,g^i,l^i)\}_{i=1}^{n}$, where $\boldsymbol{\mu^i},\alpha_i,g_i,l_i$ are the center, learning rate, goodness score and label of the prototype $p_i$. Mathematically, we modified (\ref{eq:lvq_update}) to make the learning rate individual to each prototype and a function of the goodness as described below:
\begin{align} 
&\boldsymbol{\mu^*_{t+1}} \leftarrow \boldsymbol{\mu^*_t} + \alpha_t^*(g^*_t) \Psi(y^*,\hat{y_t}) \boldsymbol{x_t}; \hspace{0.5cm} \alpha^*_t = \frac{1}{g^*_t}, \label{eq:clp_update_2}\\
&g^*_{t+1} \leftarrow \max(1,g^*_t + \Psi(y^*,\hat{y_t})). \label{eq:clp_update_3}
\end{align}

Specifically, $g^*(t)$ is the goodness score of the winner neuron at time $t$. Note that $\forall k \in [1,n], g^k_{t=0}=1$ and hence $\alpha^k_{t=0}=1$. This means in online learning, the initial input represented by a prototype serves as its starting center. Therefore, each prototype is assigned an initial learning rate of 1, indicating that it memorizes the characteristics of the initial input feature vector that led to its allocation. This becomes crucial when we discuss novelty detection in Sec.~\ref{sec:openWorld}, as encounters with novel instances trigger the allocation of new prototypes.

\subsection{Open-world Recognition and Semi-supervised Learning} \label{sec:openWorld}
CLP's another key feature is its novelty detection mechanism, which opens CLP to the possibility of encountering unknown instances. In a real-world setting, the recognition tasks are almost always open-set, i.e., the learning system will encounter classes that it has not been trained on before~\cite{DeRosa2016openMetric, Joseph2021openDetect, bohus2022wild}. CLP performs open-set learning by detecting unknown instances. As each prototype has a recognition boundary defined by a similarity threshold $\tau$, there will be regions in feature space, called ``open space'', that are not covered by any of the learned prototypes ~\cite{Bendale2015nno}. Using its novelty detection mechanism, CLP detects any sample falling into this space as a novel/unknown instance. This can be described by the CLP's novelty detector function $\nu(\boldsymbol{x}): \mathbb{R}^d \rightarrow [0,1]$ and if such novelty is detected, the model predicts label 0:
\vspace{-5pt}
\begin{equation} \label{eq:novelty_det}
\nu(\boldsymbol{x}) =  
\begin{cases}
  1 & \text{if } \forall p_i \in P, s(\boldsymbol{\mu_i},\boldsymbol{x}) < \tau \\
  0 & \text{otherwise}.
\end{cases}
\end{equation}

\begin{equation}
y^* =  
\begin{cases}
  0 & \text{if } \nu(\boldsymbol{x})=1 \\ \nonumber
  l^* & \text{otherwise},
\end{cases}
\end{equation}
where $l^*$ is the label of the prototype that is most similar to the input $x$. Subsequently, an autonomous learning system should also learn these novel instances on the fly without supervision. CLP learns such a novel instance by allocating a prototype and memorizing this sample as its center. Crucially, each novel prototype learns the center of the cluster it situates, without supervision, if some similar samples from that cluster are encountered. In the beginning, CLP assigns to such a prototype a unique pseudo-label $l_u \in \mathbb{N}_-$, where $\mathbb{N}_-$ is the set of negative integers. When a labeled sample $(x_t, y_t)$ is recognized by the prototype $p^j$ at time $t$, CLP updates the label of this prototype to the actual label: $l^j \leftarrow y_t$. This kind of labeling can be spontaneous or structured. One way such structured semi-supervised learning can be implemented is by allowing the learning agent to store the raw samples that triggered new prototype allocations and later ask a human user or multi-modal large language model labeler to provide labels for these stored samples, which in turn can be assigned to the corresponding prototypes. Thus, by combining novelty detection, unsupervised continual adaptation, and subsequent labeling, CLP achieves semi-supervised continual learning in an open-set setting.

Another implication of the unsupervised adaptation is the compensation for the possible concept drifts. The non-stationary nature of the real world means that the statistical properties of input data can gradually change over time, a phenomenon known as concept drift~\cite{gama2014conceptDrift}. To manage such cases, CLP keeps a small baseline plasticity level even for the most consolidated prototypes, allowing them to compensate for the gradual concept drift without supervision. 

\subsection{Learning Multi-modal Representations} \label{sec:multimodal}
Gradually drifting classes are not the only challenge in the online open-set continual learning setting. There can also be abrupt changes or unfamiliar instances of a known class. Most incremental learning methods assume the uni-modality of classes in embedding space; however, that is not guaranteed. As we will see in~Sec.~\ref{sec:experiments}, that assumption does not hold for classes encountered in natural settings that an autonomous agent is exposed to. Therefore, we designed CLP to learn each class as a set of clusters. Thanks to its novelty detection mechanism, CLP can learn simple (uni-modal) and complex (multi-modal) classes by assigning the prototypes to the classes on demand. This way, CLP can learn more complex decision boundaries than other non-parametric approaches~\cite{Mensink2013ncm,snell2017protoNet, Bendale2015nno}, which only learn a single prototype per class. In CLP's case, each class may have a different number of prototypes, allocated progressively as the learning continues. Only the maximum size of the prototype population, shared among all classes, must be specified in the beginning. For some classes, one prototype may be enough, while other more complex ones may require the allocation of more prototypes. The flexibility of this on-demand prototype allocation also means the resources are used more efficiently.

\section{Experiments}
\label{sec:experiments}

\subsection{General Experimental Setting}
To evaluate our method, we use OpenLORIS\cite{she2020openloris} dataset, collected in a real-world setting by a camera attached to a robot. The dataset includes 121 object instances divided into 40 unique classes. Every class has a varying number of object instances between 1 and 9. Each object instance is recorded considering four different environmental factors: clutter of the scene, illumination, occlusion, and pixel size of objects as shown in~Fig.~\ref{fig:clp_diagram}. For each variation, three levels of difficulty are considered. In addition, the objects are recorded in three different contexts (home, office, and mall). Hence, $4\times3\times3=36$ videos are recorded for each of the 121 objects.
We run three experiments that share the following characteristics: first, the learning is always online, i.e., the network learns one sample at a time, and second, we go through data only once, as in stream learning. We Leverage the EfficientNet-B0 architecture, specifically designed for edge devices, as the static feature extractor $G(\cdot)$, which aligns with the findings from Hayes and Kanan's online continual learning experiments~\cite{hayes2022online}. The backbone is initially trained on the ImageNet-1k dataset using supervised learning. Note that experimenting with different backbones is beyond the scope of our work.

\subsection{Fully Supervised Online Continual Learning} \label{sec:exp1}
\begin{figure}[t]
  \centering
   \includegraphics[width=0.92\linewidth]{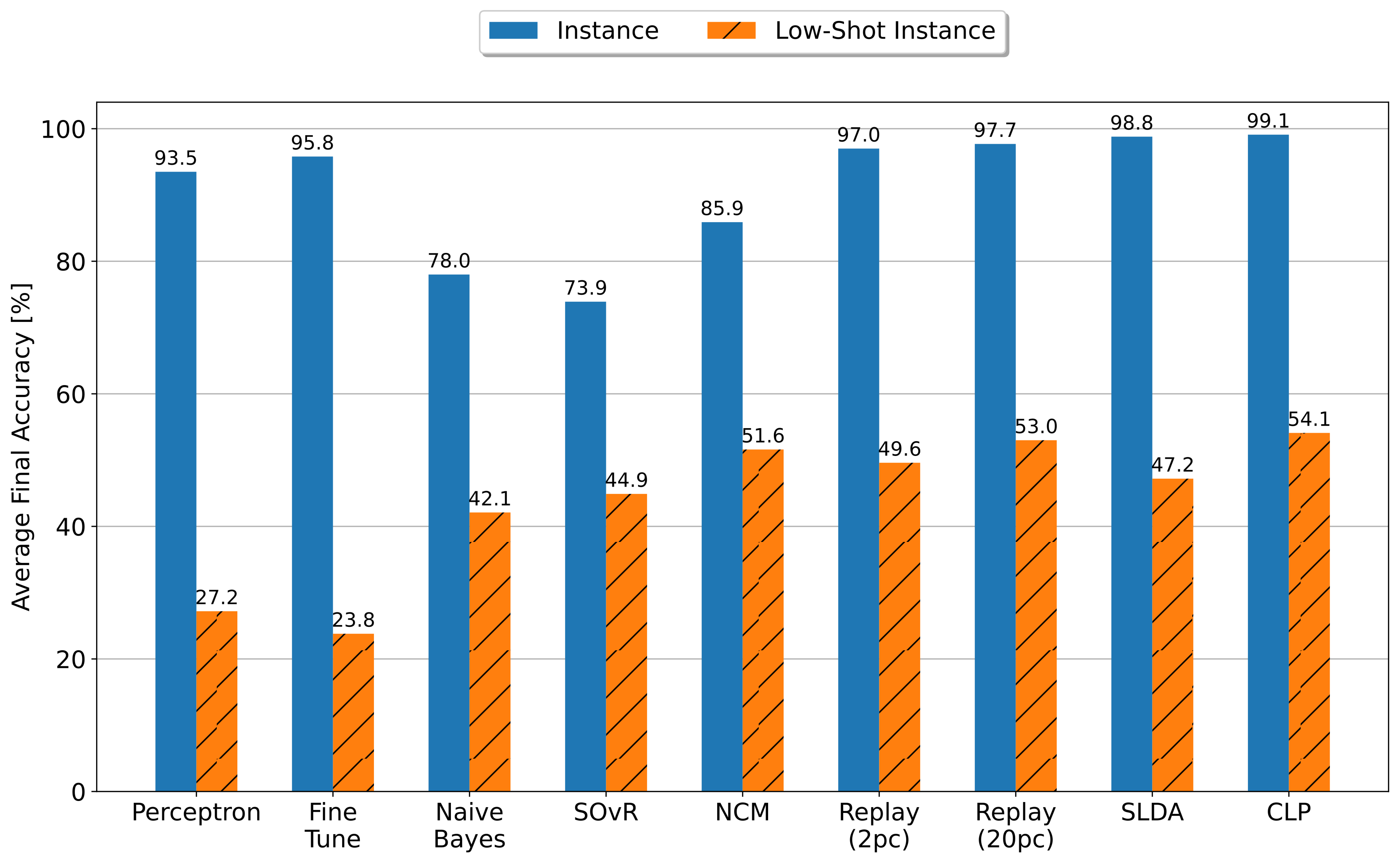}

   \caption{Fully supervised online continual learning of all classes in the OpenLORIS dataset.}
   \label{fig:res_supervised}
   \vspace{-9pt}
\end{figure}

In our first experiment, we use a protocol employed by Hayes and Kanan~\cite{hayes2022online}, i.e., fully supervised online continual learning of all classes in a single pass through the dataset. We differentiate two cases where training can be done with one or all training videos per class, respectively called ``low-shot instance" and ``instance" ordering~\cite{hayes2022online}. Using the same experimental setting, we compare CLP to the competing single-layer continual learning methods described in~\cite{hayes2022online}. We measure CLP's performance and reproduce the results of the other methods using the author's open-source code~\cite{hayes2022online}. Similarly, we ran the experiment three times and compared the results of the CLP to the other methods, as shown in Fig.~\ref{fig:res_supervised}. CLP surpasses all previous methods, both on instance and low-shot instance ordering. Therefore, CLP achieves state-of-art results for few-shot and full-shot online continual learning of the object classes on the OpenLORIS dataset. 
Note that the accuracy results for other methods are those with the EfficientNet-B0 backbone, taken from~\cite{hayes2022online}.  

\begin{figure}[t]
  \centering
   \includegraphics[width=1.\linewidth]{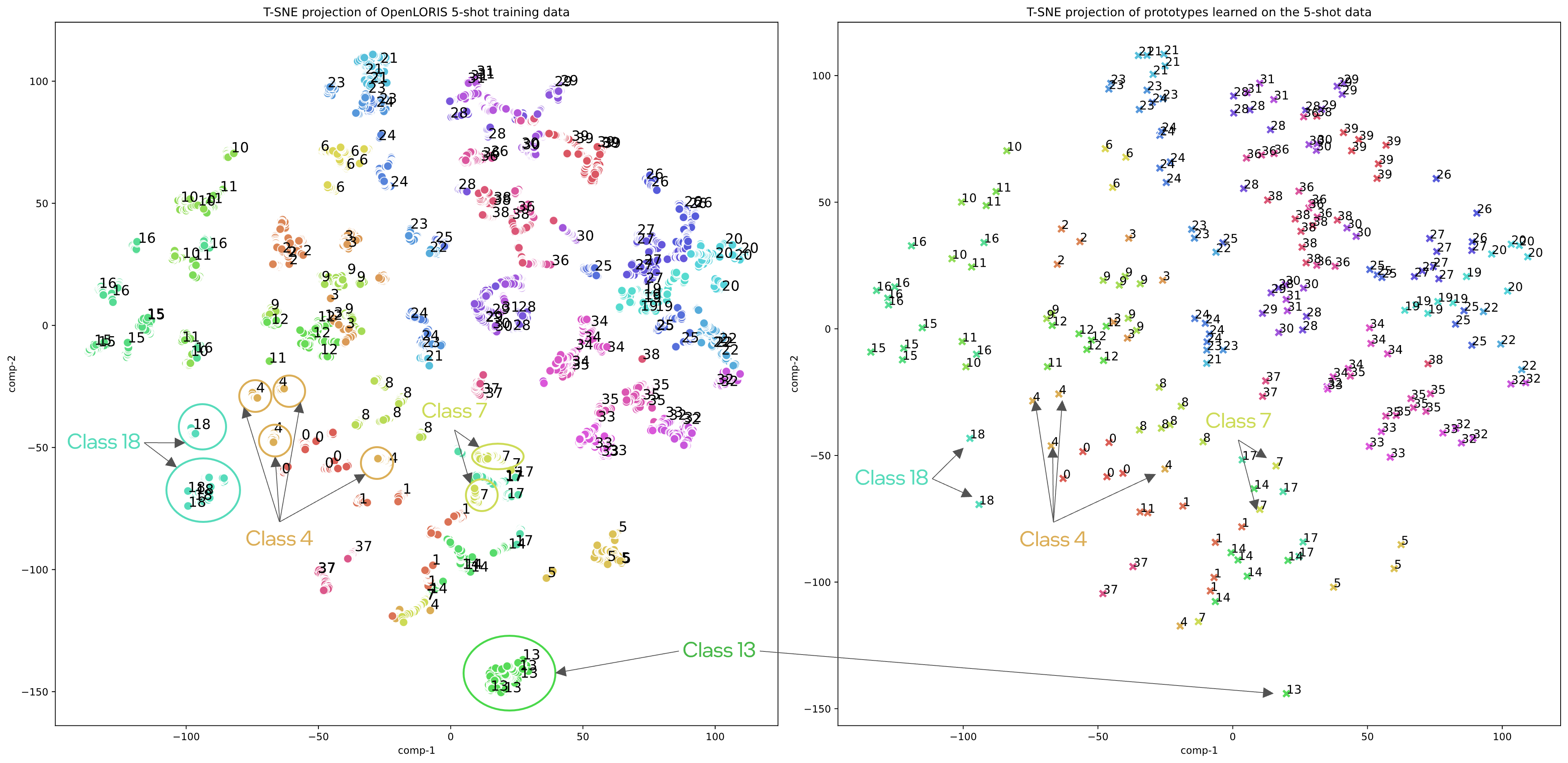}

   \caption{T-SNE visualization of the OpenLORIS features extracted by pre-trained EfficientNet-B0 backbone and learned prototypes. We randomly chose ten videos (60 frames each) for all 40 classes. The videos from each category may include different object instances but also variations of the same instances. As pointed out in the figure, some classes are represented with a single cluster and hence a single prototype (e.g., class 13), while others are clustered into varying numbers of clusters (e.g., class 4, 7, and 18) and accurately represented by multiple prototypes. This demonstrates that the methods that learn each class with a single prototype are inadequate. Conversely, CLP has adequate representational power thanks to its per-class, on-demand, multi-prototype learning mechanism.}
   \label{fig:features}
   \vspace{-12pt}
\end{figure}

\begin{figure*}[t] 
    \centering
    \begin{subfigure}{0.32\textwidth} 
        \centering
        \includegraphics[width=\linewidth]{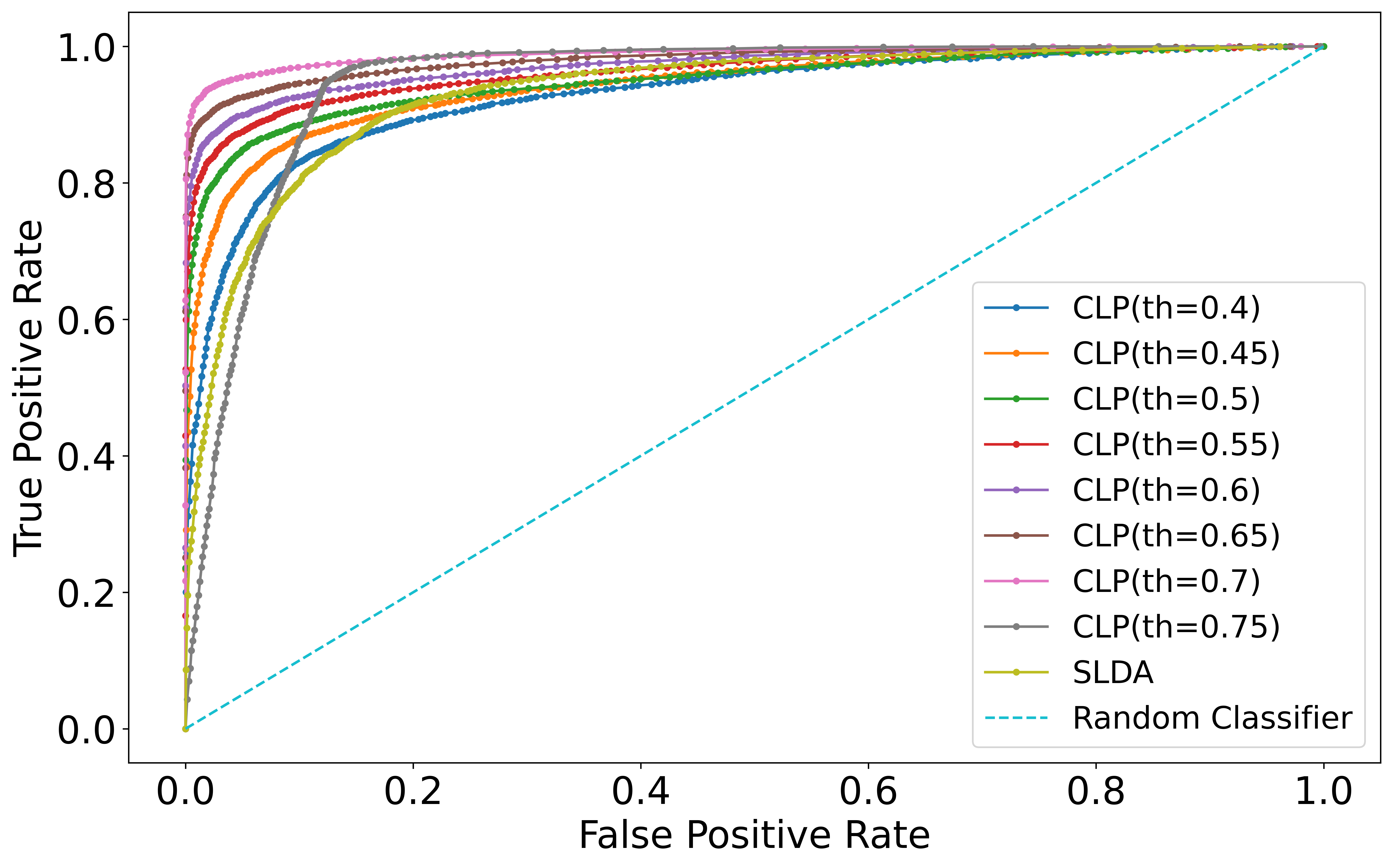} 
        \caption{} 
        \label{subfig:roc_novelty}
    \end{subfigure}
    \hfill
    \begin{subfigure}{0.32\textwidth} 
        \centering
        \includegraphics[width=\linewidth]{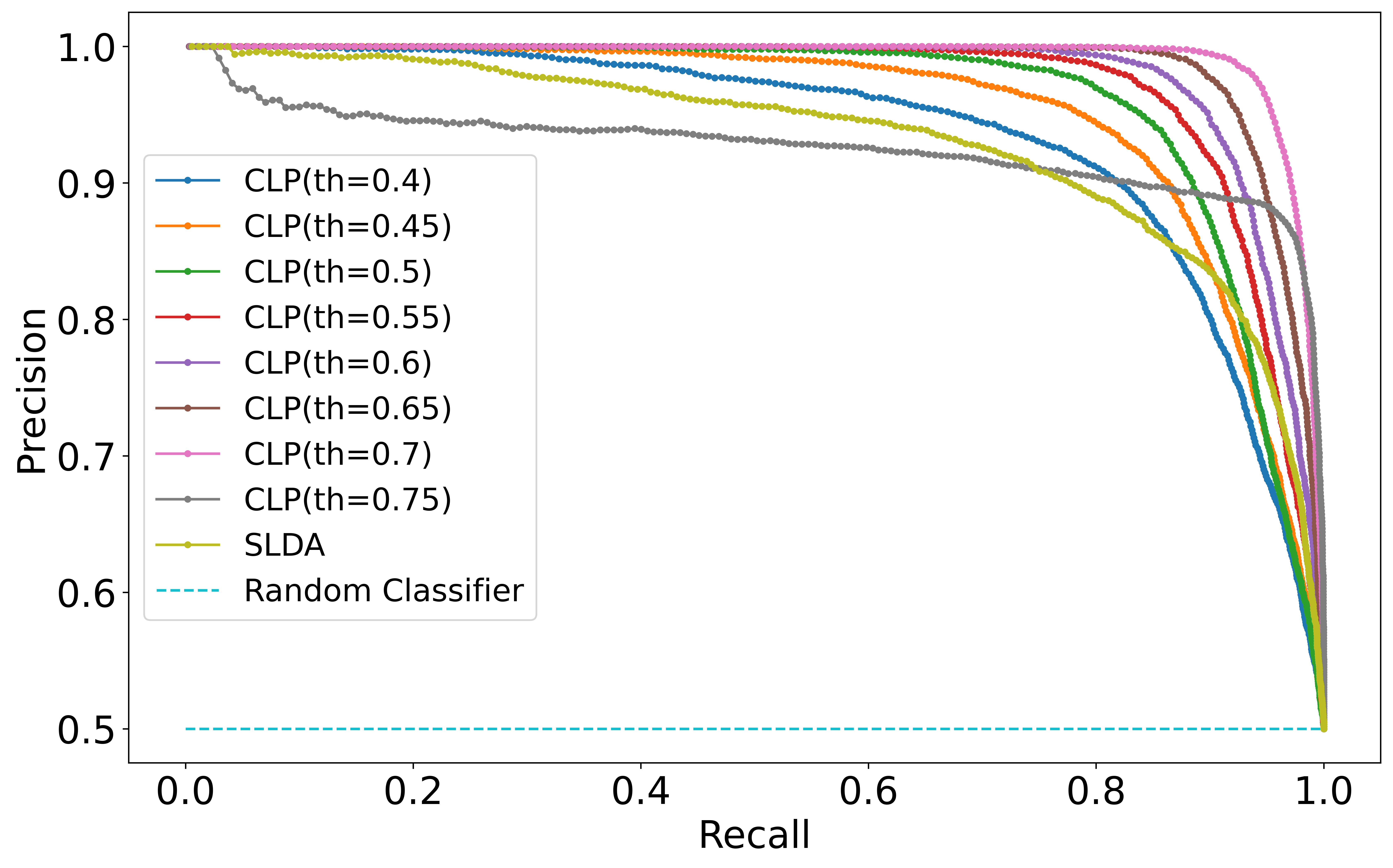} 
        \caption{} 
        \label{subfig:pr_novelty}
    \end{subfigure}
    \hfill
    \begin{subfigure}{0.32\textwidth} 
        \centering
        \includegraphics[width=\linewidth]{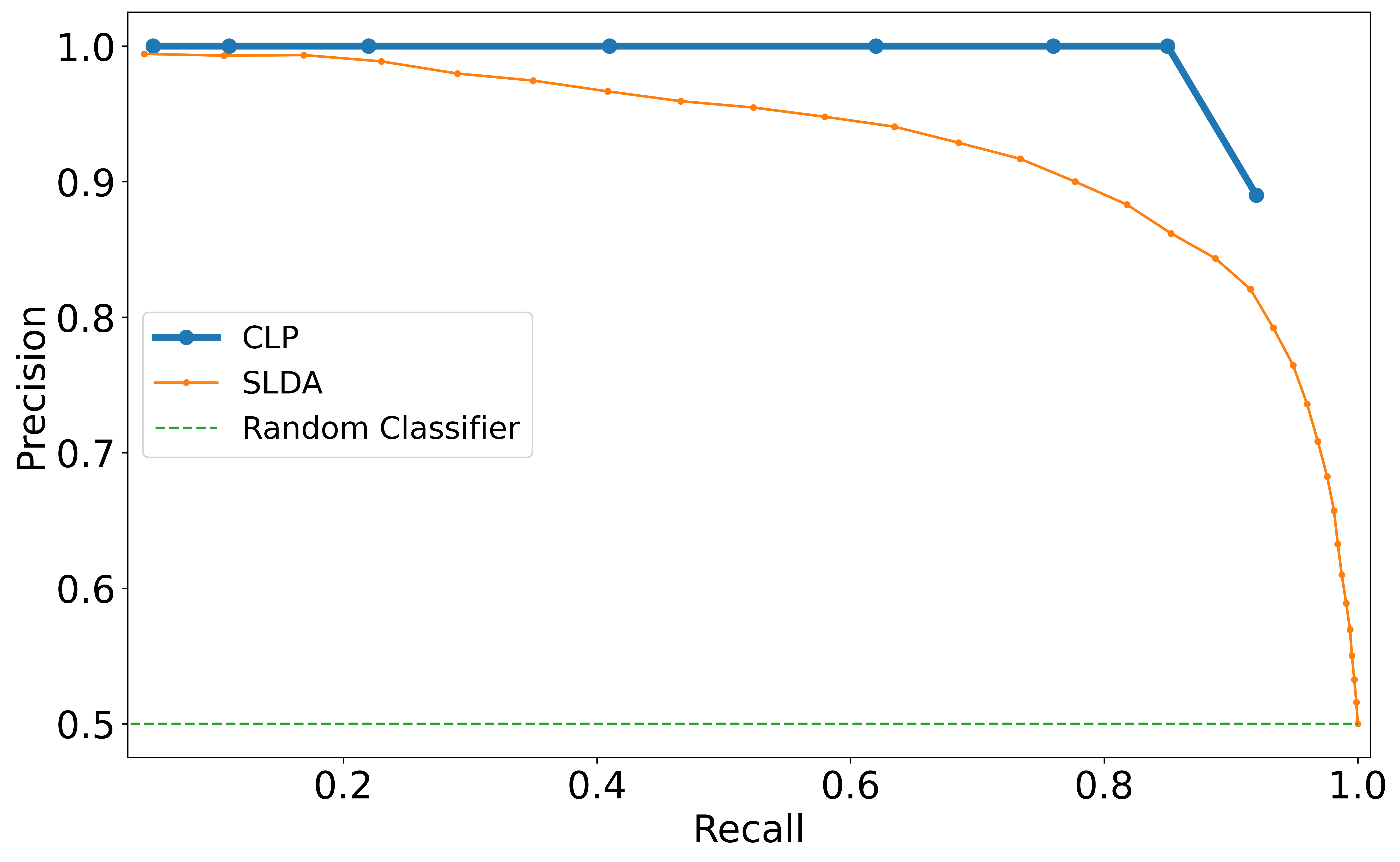} 
        \caption{} 
        \label{subfig:pr2_novelty}
    \end{subfigure}
    \caption{Performance analysis of CLP's novelty detection mechanism for open-set recognition: (a) ROC curve, (b) precision-recall curve, (c) precision-recall curve with persistent threshold}
    \label{fig:novelty_det}
    \vspace{-10pt}
\end{figure*}

\subsection{Open-set Recognition with Novelty Detection} \label{sec:res_exp2}
In the next experiment, we evaluate the CLP's open-set learning capability described in Sec.~\ref{sec:openWorld}. We divide our training set into base and novel class groups, each with 20 non-overlapping object categories. First, CLP learns the base classes using all the training samples as in the first experiment. Then, we present 8000 samples each from the test set of base and novel classes to CLP and record the output of the novelty detection function $\nu(\boldsymbol{x})$. The true labels of the novelty detection are zeros for the base and ones for the novel class samples. For each sample, we record the value of the top similarity among the prototypes as a non-thresholded measure of known class detection. Because novelty detection is regarded as a binary classifier, we can use the receiver operating characteristic curve (ROC) to illustrate its performance as the discrimination threshold is varied~\cite{fawcett2006roc}. Moreover, the area under ROC (AUROC) is a commonly used metric for measuring open-set detection capabilities~\cite{ hendrycks2016auroc1, liang2017auroc2}. We ran this experiment for CLP with different prototype thresholds $\tau$ and compared it to a version of SLDA~\cite{roady2020ood} that uses the Mahalanobis~\cite{lee2018mahalanobis} distance measure to detect novel instances. Note that $\tau$ is a CLP hyperparameter and different from the varying discrimination thresholds used for ROC analysis. The resulting ROC curves are shown in Fig.~\ref{subfig:roc_novelty}, and the corresponding AUROC values can be found in Tab.~\ref{tab:novel_det}. Undoubtedly, CLP shows superior novelty detection performance for all $\tau$'s we tested.

To further analyze the novelty detection performance, we calculated the precision and recall of the novelty detection for varying discrimination thresholds between known and unknown detection as illustrated in~Fig.~\ref{subfig:pr_novelty}. Similar to the AUROC metric, one can find the area under the precision-recall curve (AUPRC): a higher score shows that the novelty detector has a low rate of false unknown detection (high precision) and detects the majority of novel instances as novel (high recall). CLP performs significantly better than SLDA, except for the version learned with $\tau=0.75$. The reason is that with such a high threshold for similarity matching, the CLP is too aggressive in detecting instances as novel, which drops the precision of novelty detection.

Thus far, we have examined varying thresholds for the novelty detection function to create ROC and precision-recall curves. However, in an autonomous continual learning setting, the novelty detection threshold $\tau$ that CLP used for learning base classes should stay the same when detecting novel instances later because in the real world, the learning and inference are always interleaved. Thus, we calculated the precision, recall, and F1-score (defined as the harmonic mean of precision and recall) of CLP's novelty detection at the same threshold that it had used for learning base classes. The resulting values are provided in~Tab.~\ref{tab:novel_det} for the $\tau$'s used in the previous experiments and drawn in~Fig.~\ref{subfig:pr2_novelty} together with the respective curve for SLDA. Evidently, CLP is still consistently superior in terms of novelty detection, as its threshold-persistent precision-recall curve is above the SLDA curve for all values. With this, we conclude that CLP not only outperforms other online continual learning methods in two different supervised close-set settings but is also significantly superior to the nearest competitor (SLDA) in detecting novelty.

\begin{table}
  \centering 
  \setlength{\tabcolsep}{2pt}
  \begin{tabular}{@{}l*{6}{c}@{}}
    \toprule
    Classifier   & AUROC          & AUPRC          & Prec. & Recall & F1   \\
    \midrule
    CLP(th=0.55) & 0.963          & 0.972          & 1.   & 0.41   & 0.58 \\
    CLP(th=0.60) & 0.972          & 0.978          & 1.   & 0.62   & 0.76 \\
    CLP(th=0.65) & 0.98           & 0.985          & 1.   & 0.76   & 0.86 \\
    CLP(th=0.70) & \textbf{0.989} & \textbf{0.991} & 1.   & 0.85   & \textbf{0.92}\\
    CLP(th=0.75) & 0.949          & 0.926          & 0.89 & 0.92   & 0.91 \\
    SLDA (Mahal) & 0.932          & 0.931          & 0.84 & 0.9    & 0.87 \\
    \bottomrule
  \end{tabular}
  \caption{Novelty detection results.}
  \label{tab:novel_det}
  \vspace{-10pt}
\end{table}

\subsection{Semi-supervised Few-shot Continual Learning}\label{sec:res_exp3}
In a real-world setting, the intuitive next step after detecting novel instances is to learn them. Thus, we designed the next scenario as the continuation of the previous experiment, where CLP continually learned 20 base classes in a streaming fashion with all available data and labels, building the base model. After the successful validation of CLP's novelty detection~(Sec.~\ref{sec:res_exp2}), this base model can now learn the other 20 novel classes with k-shots ($k=[1,5,10,25]$) per class where each shot is a short video rather than a frame. Crucially, CLP learns these on top of its existing knowledge without supervision. Every time CLP detects a novel instance, it will allocate a new prototype and memorize this sample as its center~(Sec.~\ref{sec:openWorld}). Later, CLP continually update unlabeled prototypes without supervision as they detect more samples within their detection boundaries and recognize them using pseudo-labels. Hence, CLP operates as a novelty detection-assisted clustering algorithm when the labels are not available. In line with the semi-supervised continual learning paradigms~\cite{Shen2020online_semi_lvq, Joseph2021openDetect, Shu2020pODN}, we assume that the autonomous agent will take a snapshot of each object that triggered novelty detection, and this will be associated with the corresponding newly allocated prototype. Once in a while, a user may provide an actual label for each new prototype allocated during the autonomous, unsupervised learning phase. Of course, this requires significantly less labeling than labeling of an equivalent dataset. To summarize, the experiment simulates a real-world autonomous learning scenario with three phases: (1) base training with supervision, (2) open-world few-shot training of novel classes without supervision, and (3) labeling of the novel prototypes. Note that in both phases 1 and 2, CLP learns online through a single pass over the data. Once all the new prototypes are labeled, we test on all base and novel classes.

\begin{figure}[t]
  \centering
   \includegraphics[width=0.9\linewidth]{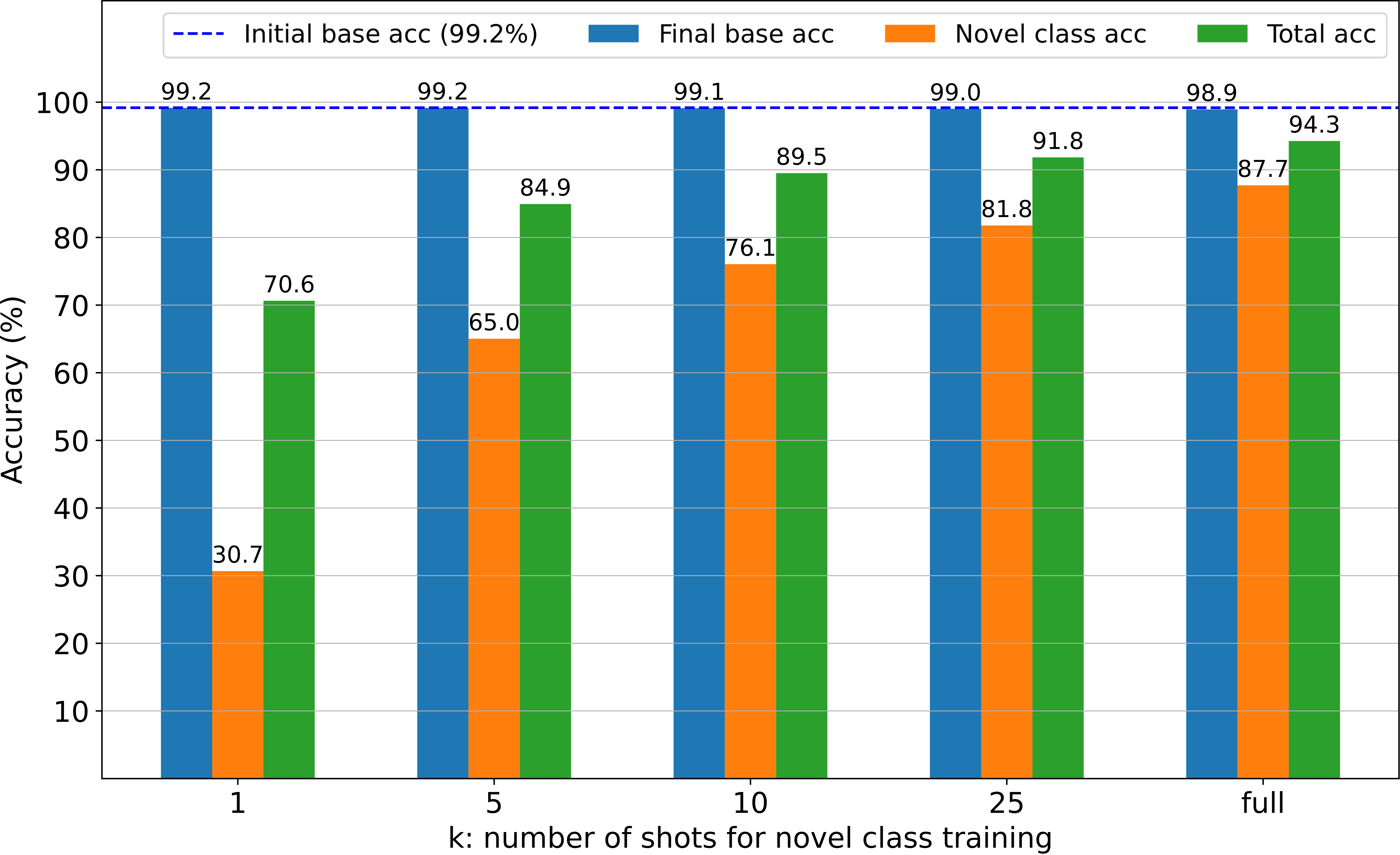}
   \caption{Few-shot semi-supervised continual learning. CLP retains base class accuracy while learning novel classes without supervision. As the number of videos provided per novel class increases, the accuracy improves significantly.}
   \label{fig:res_semi_supervised}
   \vspace{-25pt}
\end{figure}

Guided by earlier experiment outcomes (Sec.~\ref{sec:res_exp2}) and hyperparameter tuning on the validation set, we adopt a novelty detection threshold of $\tau=0.7$. Newly learned prototypes inherit the labels of their allocating instances. Post-initial training, CLP lacks exposure to base class instances, so it cannot rehearse. We record the accuracy on base classes as the baseline to later quantify forgetting. After phase 3 of learning, we measure accuracy both on base and novel classes as presented in~Fig.~\ref{fig:res_semi_supervised}. Notably, CLP maintains exceptional recognition of the base classes, exhibiting only marginal decreases in accuracy (up to a maximum of 0.3\% when employing the entire training data for novel classes). CLP also achieves unsupervised learning even with one shot. As anticipated in the context of unsupervised learning, the accuracy of novel classes is naturally lower than its supervised counterpart, e.g., in the case of 1-shot learning, we get 30.7\% accuracy compared to 54.1\% in a supervised setting~(Fig.~\ref{fig:res_supervised}). However, this accuracy improves significantly as we increase the number of shots. Notably, in the instance of the full-shot scenario, achieving a novel class accuracy of 87.7\% closely aligns with the novelty detection recall (85\%) at $\tau=0.7$. In addition, CLP allocated 1692 prototypes for base classes and, respectively, 54, 193, 292, and 602 prototypes for 1, 5, 10, and full-shot unsupervised training of novel classes.
To summarize, we demonstrate that CLP reliably learns novel classes in a few shots and without supervision, thanks to its superior novelty detection, while still remembering the base classes with minimal loss of accuracy. We hope that the CLP as a baseline for the OWCL setting will be a strong basis for future research into methods addressing similar challenges.

\section{Conclusion \& Outlook}
\label{sec:conclusion}

In our pursuit to enable truly autonomous robots that mimic the intricate learning capabilities inherent in humans, we proposed a prototype-based continual online learning algorithm with few-shot, open-world, and semi-supervised learning capabilities. 
Our flexible framework, termed Continual Learning Prototypes (CLP), showcases the capability to generate multi-modal feature representations of data and detect novel instances. Most importantly, CLP introduces a dynamic metaplasticity mechanism that provides a localized learning approach suitable for accelerators like the Loihi 2 neuromorphic chip.
Utilizing the OpenLORIS dataset, we simulate an autonomous agent that first learns a set of base classes in a supervised environment. Then, CLP is deployed into the open world, exposing the algorithm to both known and unknown categories. A key aspect of our study is demonstrating CLP's proficiency in acquiring knowledge about new classes without forgetting previously learned ones. We methodically analyze each phase of this learning process, assessing the performance at every stage. 
In the supervised mode, CLP outperforms previous state-of-the-works by producing 99.1\% and 54.1\% total accuracy when trained with all and one video per class, respectively. In an open-world setting, CLP detects novelty with high precision and recall. Moreover, CLP is also adept at learning new prototypes of these newly identified categories without supervision. Simultaneously, it dynamically adjusts the prototypes of known classes to accommodate conceptual shifts, thus producing 99\% base class and 76\% (10-shot) novel class accuracy. This multifaceted approach highlights CLP's robustness and adaptability as a learning algorithm in complex, real-world settings.

In the future, we aim to pursue these research directions: 1) validating CLP's capability to learn the variances/thresholds of the prototypes; 2) integration of object detection with CLP to build continual object detection and learning system; 3) designing a fully neuromorphic version of the algorithm running on Intel's Neuromorphic research chip Loihi 2~\cite{Davies2018}.

\bibliographystyle{IEEEtran}
\bibliography{IEEEabrv,main}

\begin{thebibliography}{10}
\providecommand{\url}[1]{#1}
\csname url@samestyle\endcsname
\providecommand{\newblock}{\relax}
\providecommand{\bibinfo}[2]{#2}
\providecommand{\BIBentrySTDinterwordspacing}{\spaceskip=0pt\relax}
\providecommand{\BIBentryALTinterwordstretchfactor}{4}
\providecommand{\BIBentryALTinterwordspacing}{\spaceskip=\fontdimen2\font plus
\BIBentryALTinterwordstretchfactor\fontdimen3\font minus \fontdimen4\font\relax}
\providecommand{\BIBforeignlanguage}[2]{{%
\expandafter\ifx\csname l@#1\endcsname\relax
\typeout{** WARNING: IEEEtran.bst: No hyphenation pattern has been}%
\typeout{** loaded for the language `#1'. Using the pattern for}%
\typeout{** the default language instead.}%
\else
\language=\csname l@#1\endcsname
\fi
#2}}
\providecommand{\BIBdecl}{\relax}
\BIBdecl

\bibitem{Hadsell2020CLreview}
R.~Hadsell, D.~Rao, A.~A. Rusu, and R.~Pascanu, ``Embracing change: Continual learning in deep neural networks,'' \emph{Trends in Cognitive Sciences}, vol.~24, pp. 1028--1040, 12 2020.

\bibitem{Yan2021der}
S.~Yan, J.~Xie, and X.~He, ``Der: Dynamically expandable representation for class incremental learning,'' \emph{CVPR}, pp. 3013--3022, 3 2021.

\bibitem{Lange2022CLreview}
M.~De~Lange, R.~Aljundi, M.~Masana, S.~Parisot, X.~Jia, A.~Leonardis, G.~Slabaugh, and T.~Tuytelaars, ``A continual learning survey: Defying forgetting in classification tasks,'' \emph{IEEE transactions on pattern analysis and machine intelligence}, vol.~44, no.~7, pp. 3366--3385, 2021.

\bibitem{hayes2022online}
T.~L. Hayes and C.~Kanan, ``Online continual learning for embedded devices,'' \emph{arXiv preprint arXiv:2203.10681}, 2022.

\bibitem{michieli2023fsocl}
U.~Michieli and M.~Ozay, ``Online continual learning for robust indoor object recognition,'' in \emph{2023 IEEE/RSJ International Conference on Intelligent Robots and Systems (IROS)}.\hskip 1em plus 0.5em minus 0.4em\relax IEEE, 2023, pp. 3849--3856.

\bibitem{Bendale2015nno}
A.~Bendale and T.~Boult, ``Towards open world recognition,'' in \emph{Proceedings of the IEEE conference on computer vision and pattern recognition}, 2015, pp. 1893--1902.

\bibitem{roady2020ood}
R.~Roady, T.~L. Hayes, R.~Kemker, A.~Gonzales, and C.~Kanan, ``Are open set classification methods effective on large-scale datasets?'' \emph{Plos one}, vol.~15, no.~9, p. e0238302, 2020.

\bibitem{jafarzadeh2020openWorldReview}
M.~Jafarzadeh, A.~R. Dhamija, S.~Cruz, C.~Li, T.~Ahmad, and T.~E. Boult, ``A review of open-world learning and steps toward open-world learning without labels,'' \emph{arXiv preprint arXiv:2011.12906}, 2020.

\bibitem{jedlicka2022metaplasticity}
P.~J. et~al, ``Contributions by metaplasticity to solving the catastrophic forgetting problem,'' \emph{Trends in Neurosciences}, vol.~45, no.~9, pp. 656--666, 2022.

\bibitem{rebuffi2017icarl}
S.-A. Rebuffi, A.~Kolesnikov, G.~Sperl, and C.~H. Lampert, ``icarl: Incremental classifier and representation learning,'' in \emph{Proceedings of the IEEE conference on Computer Vision and Pattern Recognition}, 2017, pp. 2001--2010.

\bibitem{de2021cope}
M.~De~Lange and T.~Tuytelaars, ``Continual prototype evolution: Learning online from non-stationary data streams,'' in \emph{Proceedings of the IEEE/CVF International Conference on Computer Vision}, 2021, pp. 8250--8259.

\bibitem{shin2017dgr}
H.~Shin, J.~K. Lee, J.~Kim, and J.~Kim, ``Continual learning with deep generative replay,'' \emph{Advances in neural information processing systems}, vol.~30, 2017.

\bibitem{chaudhry2018agem}
A.~Chaudhry, M.~Ranzato, M.~Rohrbach, and M.~Elhoseiny, ``Efficient lifelong learning with a-gem,'' \emph{arXiv preprint arXiv:1812.00420}, 2018.

\bibitem{kirkpatrick2017ewc}
J.~Kirkpatrick, R.~Pascanu, N.~Rabinowitz, J.~Veness, G.~Desjardins, A.~A. Rusu, K.~Milan, J.~Quan, T.~Ramalho, A.~Grabska-Barwinska \emph{et~al.}, ``Overcoming catastrophic forgetting in neural networks,'' \emph{Proceedings of the national academy of sciences}, vol. 114, no.~13, pp. 3521--3526, 2017.

\bibitem{aljundi2018mas}
R.~Aljundi, F.~Babiloni, M.~Elhoseiny, M.~Rohrbach, and T.~Tuytelaars, ``Memory aware synapses: Learning what (not) to forget,'' in \emph{Proceedings of the European conference on computer vision (ECCV)}, 2018, pp. 139--154.

\bibitem{zhang2020dmc}
J.~Zhang, J.~Zhang, S.~Ghosh, D.~Li, S.~Tasci, L.~Heck, H.~Zhang, and C.-C.~J. Kuo, ``Class-incremental learning via deep model consolidation,'' in \emph{Proceedings of the IEEE/CVF Winter Conference on Applications of Computer Vision}, 2020, pp. 1131--1140.

\bibitem{rusu2016pnn}
A.~A. Rusu, N.~C. Rabinowitz, G.~Desjardins, H.~Soyer, J.~Kirkpatrick, K.~Kavukcuoglu, R.~Pascanu, and R.~Hadsell, ``Progressive neural networks,'' \emph{arXiv preprint arXiv:1606.04671}, 2016.

\bibitem{mallya2018piggyback}
A.~Mallya, D.~Davis, and S.~Lazebnik, ``Piggyback: Adapting a single network to multiple tasks by learning to mask weights,'' in \emph{Proceedings of the European Conference on Computer Vision (ECCV)}, 2018.

\bibitem{mallya2018packnet}
A.~Mallya and S.~Lazebnik, ``Packnet: Adding multiple tasks to a single network by iterative pruning,'' in \emph{Proceedings of the IEEE conference on Computer Vision and Pattern Recognition}, 2018, pp. 7765--7773.

\bibitem{kohonen1990som}
T.~Kohonen, ``The self-organizing map,'' \emph{Proceedings of the IEEE}, vol.~78, no.~9, pp. 1464--1480, 1990.

\bibitem{losing2018incrLearningReview}
V.~Losing, B.~Hammer, and H.~Wersing, ``Incremental on-line learning: A review and comparison of state of the art algorithms,'' \emph{Neurocomputing}, vol. 275, pp. 1261--1274, 2018.

\bibitem{xu2012ilvq}
Y.~Xu, F.~Shen, and J.~Zhao, ``An incremental learning vector quantization algorithm for pattern classification,'' \emph{Neural Computing and Applications}, vol.~21, pp. 1205--1215, 2012.

\bibitem{beyer2013dyng}
O.~Beyer and P.~Cimiano, ``Dyng: Dynamic online growing neural gas for stream data classification.'' in \emph{ESANN}, 2013.

\bibitem{parisi2018prototypeSequence}
G.~I. Parisi, J.~Tani, C.~Weber, and S.~Wermter, ``Lifelong learning of spatiotemporal representations with dual-memory recurrent self-organization,'' \emph{Frontiers in neurorobotics}, vol.~12, p.~78, 2018.

\bibitem{zhu2021protoAug}
F.~Zhu, X.-Y. Zhang, C.~Wang, F.~Yin, and C.-L. Liu, ``Prototype augmentation and self-supervision for incremental learning,'' in \emph{Proceedings of the IEEE/CVF Conference on Computer Vision and Pattern Recognition}, 2021, pp. 5871--5880.

\bibitem{hayes2020deepSLDA}
T.~L. Hayes and C.~Kanan, ``Lifelong machine learning with deep streaming linear discriminant analysis,'' in \emph{Proceedings of the IEEE/CVF conference on computer vision and pattern recognition workshops}, 2020, pp. 220--221.

\bibitem{hayes2019protoReplay}
T.~L. Hayes, N.~D. Cahill, and C.~Kanan, ``Memory efficient experience replay for streaming learning,'' in \emph{2019 International Conference on Robotics and Automation (ICRA)}.\hskip 1em plus 0.5em minus 0.4em\relax IEEE, 2019, pp. 9769--9776.

\bibitem{snell2017protoNet}
J.~Snell, K.~Swersky, and R.~Zemel, ``Prototypical networks for few-shot learning,'' \emph{Advances in neural information processing systems}, vol.~30, 2017.

\bibitem{hersche2022constrained}
M.~Hersche, G.~Karunaratne, G.~Cherubini, L.~Benini, A.~Sebastian, and A.~Rahimi, ``Constrained few-shot class-incremental learning,'' in \emph{Proceedings of the IEEE/CVF Conference on Computer Vision and Pattern Recognition}, 2022, pp. 9057--9067.

\bibitem{allen2019imp}
K.~R.~A. et~al, ``Infinite mixture prototypes for few-shot learning,'' \emph{36th International Conference on Machine Learning, ICML 2019}, vol. 2019-June, pp. 348--357, 2 2019.

\bibitem{ayub2020fscilIROS}
A.~Ayub and A.~R. Wagner, ``Tell me what this is: Few-shot incremental object learning by a robot,'' in \emph{2020 IEEE/RSJ International Conference on Intelligent Robots and Systems (IROS)}.\hskip 1em plus 0.5em minus 0.4em\relax IEEE, 2020, pp. 8344--8350.

\bibitem{ayub2020cbcl}
------, ``Cognitively-inspired model for incremental learning using a few examples,'' in \emph{Proceedings of the IEEE/CVF Conference on Computer Vision and Pattern Recognition Workshops}, 2020, pp. 222--223.

\bibitem{Shen2020online_semi_lvq}
Y.~Y. Shen, Y.~M. Zhang, X.~Y. Zhang, and C.~L. Liu, ``Online semi-supervised learning with learning vector quantization,'' \emph{Neurocomputing}, vol. 399, pp. 467--478, 7 2020.

\bibitem{bohus2022wild}
D.~Bohus, S.~Andrist, A.~Feniello, N.~Saw, and E.~Horvitz, ``Continual learning about objects in the wild: An interactive approach,'' in \emph{Proceedings of the 2022 International Conference on Multimodal Interaction}, 2022, pp. 476--486.

\bibitem{Ren2021wander}
M.~Ren, M.~L. Iuzzolino, M.~C. Mozer, and R.~S. Zemel, ``Wandering within a world: Online contextualized few-shot learning,'' \emph{arXiv preprint arXiv:2007.04546}, 2020.

\bibitem{Joseph2021openDetect}
K.~Joseph, S.~Khan, F.~S. Khan, and V.~N. Balasubramanian, ``Towards open world object detection,'' in \emph{Proceedings of the IEEE/CVF conference on computer vision and pattern recognition}, 2021, pp. 5830--5840.

\bibitem{Kudithipudi2022bioCL}
D.~Kudithipudi, M.~Aguilar-Simon, J.~Babb, M.~Bazhenov, D.~Blackiston, J.~Bongard, A.~P. Brna, S.~Chakravarthi~Raja, N.~Cheney, J.~Clune \emph{et~al.}, ``Biological underpinnings for lifelong learning machines,'' \emph{Nature Machine Intelligence}, vol.~4, no.~3, pp. 196--210, 2022.

\bibitem{Biehl2016}
M.~Biehl, B.~Hammer, and T.~Villmann, ``Prototype-based models in machine learning,'' \emph{Wiley Interdisciplinary Reviews: Cognitive Science}, vol.~7, pp. 92--111, 3 2016.

\bibitem{hou2019unifiedClassifier}
S.~Hou, X.~Pan, C.~C. Loy, Z.~Wang, and D.~Lin, ``Learning a unified classifier incrementally via rebalancing,'' in \emph{Proceedings of the IEEE/CVF conference on computer vision and pattern recognition}, 2019, pp. 831--839.

\bibitem{Davies2018}
M.~Davies, N.~Srinivasa, T.~H. Lin, G.~Chinya, Y.~Cao, S.~H. Choday, G.~Dimou, P.~Joshi, N.~Imam, S.~Jain, Y.~Liao, C.~K. Lin, A.~Lines, R.~Liu, D.~Mathaikutty, S.~McCoy, A.~Paul, J.~Tse, G.~Venkataramanan, Y.~H. Weng, A.~Wild, Y.~Yang, and H.~Wang, ``Loihi: A neuromorphic manycore processor with on-chip learning,'' \emph{IEEE Micro}, vol.~38, pp. 82--99, 2018.

\bibitem{hajizada2022interactive}
E.~Hajizada, P.~Berggold, M.~Iacono, A.~Glover, and Y.~Sandamirskaya, ``Interactive continual learning for robots: a neuromorphic approach,'' in \emph{Proceedings of the International Conference on Neuromorphic Systems 2022}, 2022, pp. 1--10.

\bibitem{hamker200individualLR}
F.~H. Hamker, ``Life-long learning cell structures—continuously learning without catastrophic interference,'' \emph{Neural Networks}, vol.~14, no. 4-5, pp. 551--573, 2001.

\bibitem{DeRosa2016openMetric}
R.~De~Rosa, T.~Mensink, and B.~Caputo, ``Online open world recognition,'' \emph{arXiv preprint arXiv:1604.02275}, 2016.

\bibitem{gama2014conceptDrift}
J.~Gama, I.~{\v{Z}}liobait{\.e}, A.~Bifet, M.~Pechenizkiy, and A.~Bouchachia, ``A survey on concept drift adaptation,'' \emph{ACM computing surveys (CSUR)}, vol.~46, no.~4, pp. 1--37, 2014.

\bibitem{Mensink2013ncm}
T.~Mensink, J.~Verbeek, F.~Perronnin, and G.~Csurka, ``Distance-based image classification: Generalizing to new classes at near-zero cost,'' \emph{IEEE Transactions on Pattern Analysis and Machine Intelligence}, vol.~35, pp. 2624--2637, 2013.

\bibitem{she2020openloris}
Q.~She, F.~Feng, X.~Hao, Q.~Yang, C.~Lan, V.~Lomonaco, X.~Shi, Z.~Wang, Y.~Guo, Y.~Zhang \emph{et~al.}, ``Openloris-object: A robotic vision dataset and benchmark for lifelong deep learning,'' in \emph{2020 IEEE international conference on robotics and automation (ICRA)}.\hskip 1em plus 0.5em minus 0.4em\relax IEEE, 2020, pp. 4767--4773.

\bibitem{fawcett2006roc}
T.~Fawcett, ``An introduction to roc analysis,'' \emph{Pattern recognition letters}, vol.~27, no.~8, pp. 861--874, 2006.

\bibitem{hendrycks2016auroc1}
D.~Hendrycks and K.~Gimpel, ``A baseline for detecting misclassified and out-of-distribution examples in neural networks,'' \emph{arXiv preprint arXiv:1610.02136}, 2016.

\bibitem{liang2017auroc2}
S.~Liang, Y.~Li, and R.~Srikant, ``Enhancing the reliability of out-of-distribution image detection in neural networks,'' \emph{arXiv preprint arXiv:1706.02690}, 2017.

\bibitem{lee2018mahalanobis}
K.~Lee, K.~Lee, H.~Lee, and J.~Shin, ``A simple unified framework for detecting out-of-distribution samples and adversarial attacks,'' \emph{Advances in neural information processing systems}, vol.~31, 2018.

\bibitem{Shu2020pODN}
Y.~Shu, Y.~Shi, Y.~Wang, T.~Huang, and Y.~Tian, ``P-odn: Prototype-based open deep network for open set recognition,'' \emph{Scientific Reports 2020 10:1}, vol.~10, pp. 1--13, 4 2020.

\end{thebibliography}

\clearpage
\onecolumn
\setcounter{page}{1}
\begin{center}
    \Large\textbf{Supplementary Material}
  \end{center}
  
\section{CLP's Pseudo-code}
\label{sec:pseudocode}
In this section, we provide the detailed pseudo-code for CLP. Firstly in Algorithm~\ref{algorithm_clp}, we express the CLP's online training, including the supervised and unsupervised portions, the novelty detection mechanism, and the labeling of the prototypes. The CLP training algorithm (Algorithm~\ref{algorithm_clp}), in turn, uses Algorithm~\ref{allocate_prototype} to allocate new prototypes and Algorithm~\ref{update_prototype} to update winner prototypes as the learning progresses, as detailed next. For the details of the CLP algorithm, please refer to Sec.~\ref{sec:clp}. We will release the source code for PyTorch implementation of CLP soon\footnote{We developed CLP independently and entirely from scratch. On the other hand, we implemented our experimental scenarios on top of the source code provided by Hayes and Kanan~\cite{hayes2022online} in the~\hyperlink{Embedded-CL}{https://github.com/tyler-hayes/Embedded-CL}, while also modifying it.}.

\begin{algorithm*}[ht]
\caption{Continually Learning Prototypes (CLP)}
\label{algorithm_clp}
\begin{algorithmic}
\REQUIRE Training data: $(\boldsymbol{x_1},y_1),...,(\boldsymbol{x_k},y_k)\in D$  \hfill $\triangleright$ a mix of labeled and unlabeled data
\REQUIRE $P = \{(\boldsymbol{\mu^i},\alpha^i,g^i,l^i)\}_{i=1}^{N}$  \hfill $\triangleright$ prototype population
\STATE Initialize $p\in P \leftarrow (\vec{0},1,1,0)$  \hfill $\triangleright$ all prototypes initialized to zero vector, learning rate of 1 and label of ``0"
\FOR{each data point $x_i$ in $D$}  
    \STATE Find the similarities between $\boldsymbol{x_i}$ and $P$, and perform novelty detection
    \IF{$\forall p_j \in P, s(\boldsymbol{\mu_j},\boldsymbol{x_i}) < \tau$}
        \STATE $Allocate(x_i, y_i)$ \hfill $\triangleright$ if all prototypes are below the similarity threshold allocate a new prototype in $P$
        \STATE \textbf{break} \hfill $\triangleright$ go to instance $i+1$
    \ENDIF
    \STATE $\triangleright$ Otherwise, there is a winner with the index of $*$
    \IF {$x_i$ has a label}
        \STATE $\triangleright$ supervised learning
        \IF {$l^{*}<0$} 
            \STATE $l^*\leftarrow y_i$ \hfill $\triangleright$ If the winner has a pseudo label, assign the actual label to this prototype
        \ENDIF
        \STATE $\Psi(l^*,y_i) =  
          \begin{cases}
            +1 & \text{if } l^*=y_i \hspace{9.36cm} \text{ $\triangleright$ correct prediction}\\
            -1 & \text{otherwise}   \hspace{9.05cm}\text{ $\triangleright$ incorrect prediction}
          \end{cases} $
        \STATE $Update(p^*, x_i, \Psi)$ \hfill $\triangleright$ update the prototype based on the evaluation of the prediction
    \ELSE 
        \STATE $\triangleright$ if $x_i$ does not have a label, perform unsupervised update
        \STATE $Update(p^*, x_i, 0.5)$ \hfill $\triangleright$  as a positive update with half of the learning rate  
    \ENDIF  
\ENDFOR
\end{algorithmic}
\end{algorithm*}

\begin{algorithm*}[ht]
\caption{Allocate: Allocating a prototype}
\label{allocate_prototype}
\begin{algorithmic}
\REQUIRE input feature vector $\boldsymbol{x_t}$, id of the unallocated prototype $u$
\REQUIRE (optionally) label of the current instance $y_t$

\STATE $\boldsymbol{\mu^u_{t+1}} \leftarrow \boldsymbol{x_t}$ \hfill $\triangleright$ update the center of the next unallocated prototype to the current input feature vector
\STATE $g^u_{t+1} \leftarrow g^u_t + 1$                                                          \hfill $\triangleright$ update the goodness score of this prototype
\STATE $\alpha^u_{t+1} = \frac{1}{g^u_{t+1}}$                                                       \hfill $\triangleright$ update the learning rate of this prototype
\STATE $\boldsymbol{\mu^u} \leftarrow \dfrac{\boldsymbol{\mu^u} }{||\boldsymbol{\mu^u} ||} $         \hfill $\triangleright$ normalize the center vector of the prototype
\IF{$x_t$ is labeled}
    \STATE $l^k \leftarrow y_t$ 
\ELSE
    \STATE $l^k \leftarrow$ {unique negative integer as a pseudo-label} 
\ENDIF
\end{algorithmic}
\end{algorithm*}

\vspace{3cm}

\begin{algorithm*}[ht]
\caption{Update: Updating a prototype}
\label{update_prototype}
\begin{algorithmic}
\REQUIRE Winner prototype $p^*$, input feature vector $\boldsymbol{x_t}$, the factor for direction and scaling of the update $\psi \in [-1,1]$
\REQUIRE (optionally) label of the current instance $y_t$
\REQUIRE $P = \{(\boldsymbol{\mu^i},\alpha^i,g^i,l^i)\}$  \hfill $\triangleright$ prototype population
\STATE $(\boldsymbol{\mu^*},\alpha^*,g^*,l^*) \leftarrow p^*$
\STATE $\boldsymbol{\mu^*_{t+1}} \leftarrow \boldsymbol{\mu^*_t} + \psi \alpha_t^*(g^*_t) \boldsymbol{x_t}$ \hfill $\triangleright$ update the center of the winner prototype
\STATE $g^*_{t+1} \leftarrow g^*_t + \psi$                                                          \hfill $\triangleright$ update the goodness score of this prototype
\STATE $\alpha^*_{t+1} = \frac{1}{g^*_{t+1}}$                                                       \hfill $\triangleright$ update the learning rate of this prototype
\STATE $\boldsymbol{\mu^*} \leftarrow \dfrac{\boldsymbol{\mu^*} }{||\boldsymbol{\mu^*} ||} $         \hfill $\triangleright$ normalize the center vector of this prototype
\IF{$\psi = -1$}
    \STATE $\triangleright$ if it is a negative update, we sort prototypes based on similarity and update all next $m$ most similar prototypes
    \FOR{$k \in [1,m]$}
        \STATE $\psi = \Psi(l^k,y_t)$                                              \hfill $\triangleright$ evaluate prediction of the next most similar prototype
        \STATE $\boldsymbol{\mu^k_{t+1}} \leftarrow \boldsymbol{\mu^k_t} + \psi \alpha_t^k(g^k_t) \boldsymbol{x_t}$   
        \STATE $g^k_{t+1} \leftarrow g^k_t + \psi$                                                          
        \STATE $\alpha^k_{t+1} = \frac{1}{g^k_{t+1}}$           
        \STATE $\boldsymbol{\mu^k} \leftarrow \dfrac{\boldsymbol{\mu^k} }{||\boldsymbol{\mu^k} ||} $
    \ENDFOR
    \IF{we don't have any positive match among $m$ most similar prototypes}
        \STATE $Allocate(x_t, y_t)$ \hfill $\triangleright$ allocate a new prototype in $P$
    \ENDIF
\ENDIF

\end{algorithmic}
\end{algorithm*}

\section{Model Size Analysis}
\label{sec:model_size}

\begin{figure*}[h!] 
    \centering
    \begin{subfigure}{0.49\textwidth} 
        \centering
        \includegraphics[width=\linewidth]{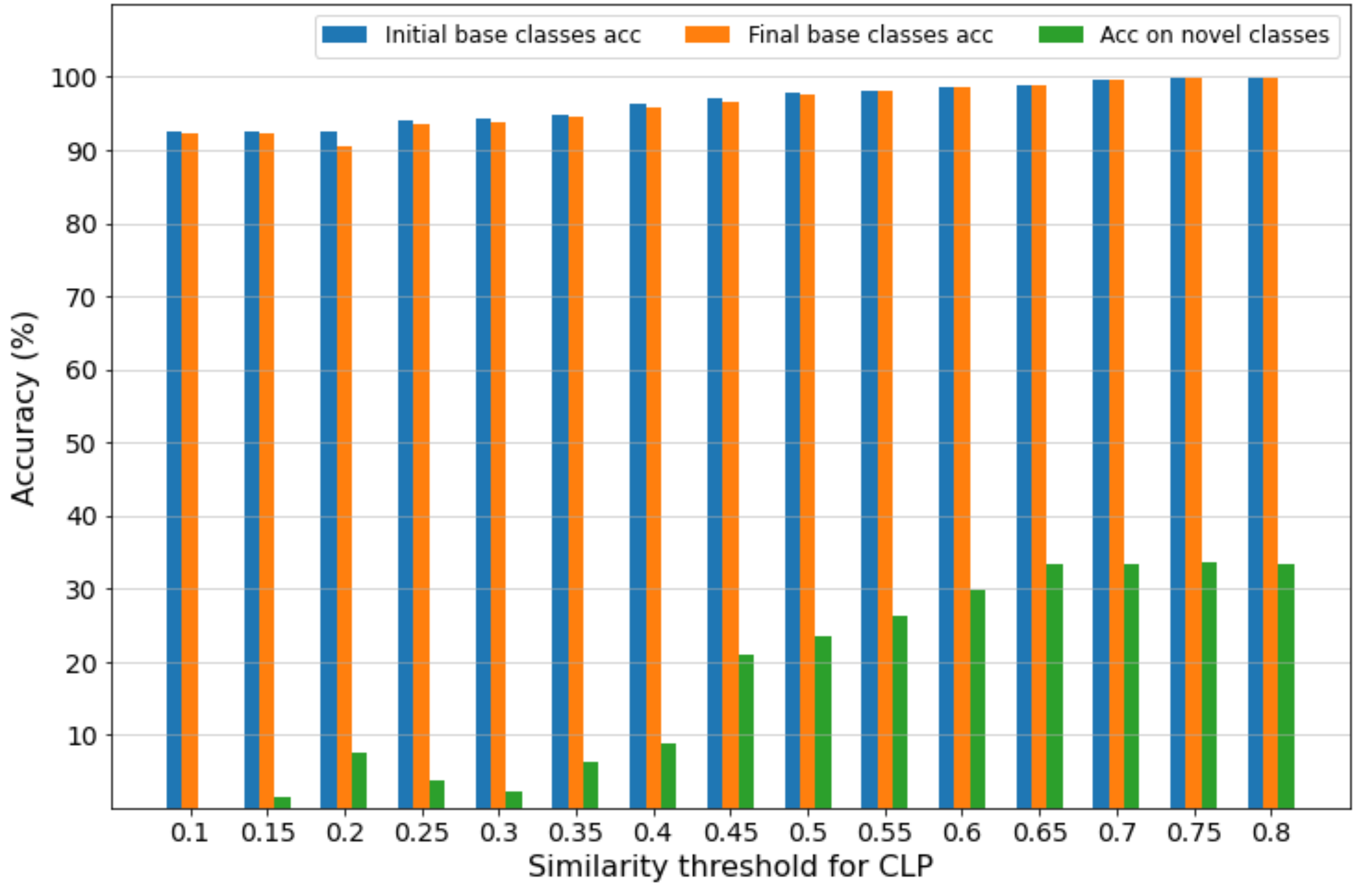} 
        \caption{} 
        \label{subfig:acc_vs_th}
    \end{subfigure}
    \hfill
    \begin{subfigure}{0.49\textwidth} 
        \centering
        \includegraphics[width=\linewidth]{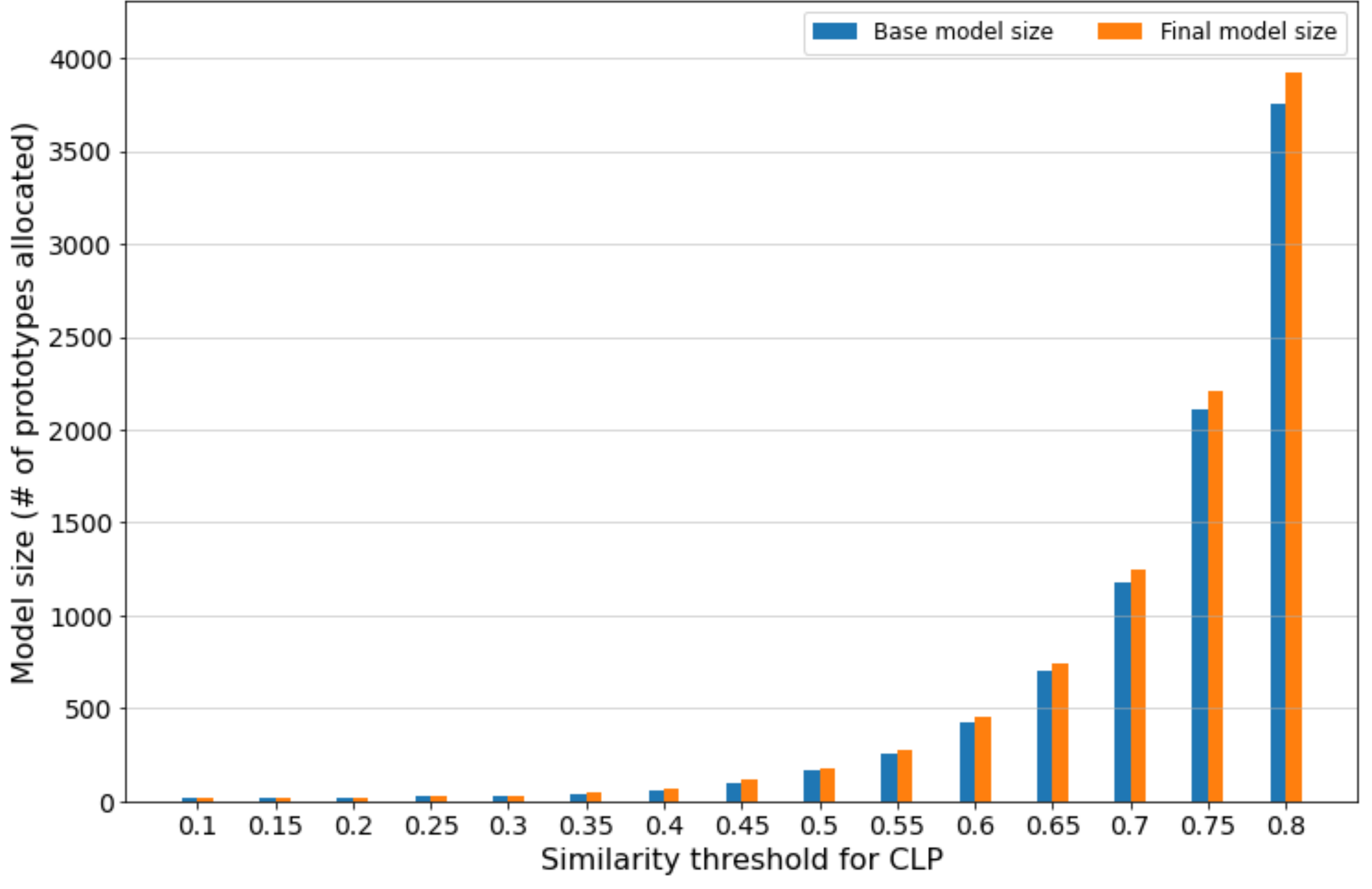} 
        \caption{} 
        \label{subfig:model_size_vs_th}
    \end{subfigure}
    \caption{Model size analysis of CLP with the varying similarity threshold $\tau$. CLP continually learned 20 base classes using all the training data with supervision and then 20 novel classes using 1-shot without supervision. (a) The accuracy of initial base class accuracy, novel class accuracy, and final base class accuracy after learning novel classes is depicted versus the similarity threshold. (b) Model size (number of the allocated prototypes) versus similarity threshold. The base model size (blue) constitutes a significant portion of the final model size, as the base classes are trained with all training data and supervision, while novel classes (orange) only take up a small portion because they are learned unsupervised with only one shot. The model size explodes when the threshold is too high while not giving any benefit in terms of accuracy. }
    \label{fig:model_size_analysis}
\end{figure*}

In the Sec.~\ref{sec:multimodal}, we described the issue of uni-modal representation of classes in a natural setting and how CLP addresses this issue with multi-modal representation using multiple prototypes per class. Crucially, CLP allocates varying numbers of prototypes per class as it continually learns classes, assigning more prototypes to the classes that are more scattered or confused in the feature space. This section will analyze model size, i.e., the number of prototypes allocated by CLP in our experiments. In the first analysis, we used the semi-supervised few-shot online continual learning scenario described in Sec.~\ref{sec:res_exp3}. We run this experiment multiple times with different similarity thresholds $\tau$ to understand the effect of this hyperparameter on different types of accuracies and the learned model size. The results are shown in Fig.~\ref{fig:model_size_analysis}. Multiple observations can be made from this figure. Firstly, as expected with high thresholds, the boundaries of the existing prototypes are tighter, causing CLP to more aggressively allocate prototypes, increasing model size Fig.~\ref{subfig:model_size_vs_th}. However, we also observed that even though larger models perform better in terms of accuracy in the beginning, we get diminishing returns, and after some point, accuracy does not go up anymore, but model size explodes. Nevertheless, our choice of the $\tau=0.7$ in the main experiments is also validated here, as shown in Fig.~\ref{subfig:acc_vs_th}, where beyond $\tau=0.7$, there is no significant increase in accuracy. Of course, one could choose $\tau=0.6$ or $0.65$ to decrease the model size, with some sacrifice in performance. Besides, we observe that for the supervised portion of the learning (blue bars in Fig.~\ref{subfig:acc_vs_th}), the models, even with the fewest number of prototypes, perform well ($>90\%$). However, as the similarity threshold is very small in these cases, the prototypes have large boundaries, covering most of the feature space, thus rendering the novelty detection incapable and subsequently crippling unsupervised learning (green bars in left half of the Fig.~\ref{subfig:acc_vs_th}).

\begin{figure}[h]
  \centering
   \includegraphics[width=0.4\linewidth]{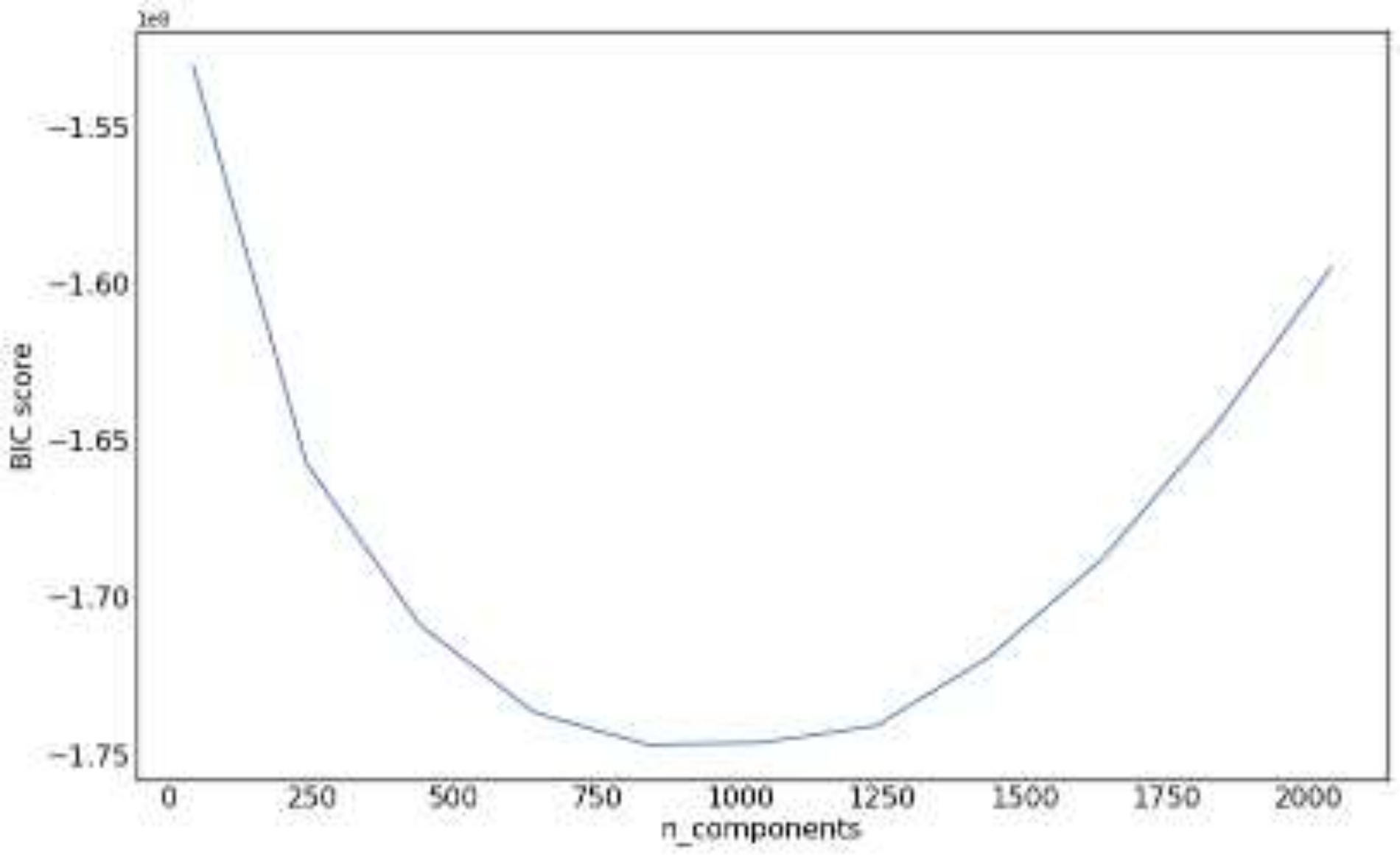}

   \caption{Bayesian Information Criterion (BIC) curve with varying number of components for Gaussian Mixture Modelling. To find most optimal number of components, we need to minimize BIC, which corresponds to 840 in this case, which is in line with the number of prototypes allocated by CLP for the same training data.}
   \label{fig:gmm_bic}
\end{figure}

In the second analysis, we compared the number of prototypes allocated by CLP to the number of the Gaussian components that gave the best results for the Gaussian Mixture Model (GMM) of the same data in terms of the Bayesian Information criterion. The Gaussian mixture model is an unsupervised probabilistic clustering method that assumes all the data points are generated from a mixture of a finite number of Gaussian distributions with unknown parameters. GMMs can compute the Bayesian Information Criterion (BIC) to assess the number of clusters in the data. Therefore, we extracted features for randomly selected ten videos from all 40 classes of the OpenLORIS dataset and modeled this with GMM using different numbers of components, between 40 and 2000. We chose the spherical covariance type, where each component has its single variance, as this would be the most similar to the prototypes of CLP. We analyzed the BIC curve (Fig.~\ref{fig:gmm_bic}) to determine the optimal number of components for GMM minimizing the BIC is around 840. This is very similar to the number of prototypes CLP allocated (846) to learn the same data in a supervised manner with $\tau=0.7$. This quantitively validates two of our hypothesizes: firstly, more than one prototype/component per class is necessary to reliably model data in a natural setting, and secondly, during continual learning, CLP allocates the near-to-optimal number of prototypes to model underlying data from the standpoint of Bayesian information criterion. Furthermore, again we validate that $\tau=0.7$ is also the best choice in terms of multi-modal representation learning. 

\end{document}